\documentclass[conference,a4paper]{IEEEtran}
\usepackage{times}
\usepackage{epsfig}
\usepackage{graphicx}
\usepackage[cmex10]{amsmath}
\usepackage{amssymb}
\usepackage{balance}
\usepackage{mathtools}
\usepackage{dsfont}
\usepackage{enumerate}
\usepackage{booktabs}
\usepackage{cite}
\usepackage{array}
\usepackage[top=0.75in, bottom=1.569in, left=0.514in, right=0.514in]{geometry}

\begin{document}

\title{Action Classification with Locality-constrained Linear Coding}

\author{\IEEEauthorblockN{Hossein Rahmani\IEEEauthorrefmark{1},
Arif Mahmood\IEEEauthorrefmark{2},
Du Huynh\IEEEauthorrefmark{3} and
Ajmal Mian\IEEEauthorrefmark{4}}
\IEEEauthorblockA{School of Computer Science and Software Engineering\\ The University of Western Australia, 35 Stirling Highway, Crawley, WA 6009 Australia}
\IEEEauthorblockA{Email: \IEEEauthorrefmark{1}hossein@csse.uwa.edu.au \IEEEauthorrefmark{2} arif.mahmood@uwa.edu.au\\ \IEEEauthorrefmark{3} du.huynh@uwa.edu.au \IEEEauthorrefmark{4} ajmal.mian@uwa.edu.au}
}

\maketitle

\begin{abstract}
  We propose an action classification algorithm which uses
  Locality-constrained Linear Coding (LLC) to capture discriminative
  information of human body variations in each spatio-temporal subsequence of
  a video sequence. Our proposed method divides the input video into equally
  spaced overlapping spatio-temporal subsequences, each of which is decomposed
  into blocks and then cells. We use the Histogram of Oriented Gradient
  (HOG3D) feature to encode the information in each cell.  We justify the use
  of LLC for encoding the block descriptor by demonstrating its superiority
  over Sparse Coding (SC). Our sequence descriptor is obtained via a logistic
  regression classifier with L2 regularization. We evaluate and compare our
  algorithm with ten state-of-the-art algorithms on five benchmark datasets.
  Experimental results show that, on average, our algorithm gives better
  accuracy than these ten algorithms. 
\end{abstract}

\section{Introduction}
Human action recognition from videos is a challenging problem. Differences in
viewing direction and distance, body sizes of the human subjects, clothing,
and style of performing the action are some of the main factors making human
action recognition difficult.  Most of the previous approaches to action
recognition have focused on using traditional RGB cameras \cite{main,
  RGB1,RGB2}. Since the release of Microsoft Kinect depth camera, depth based
human action recognition methods
~\cite{MyWACV14,DSTIP,Wang2012,HON4D,ActionLet2012,DMM,Bag3DPoints,
  Hand-Pose-Estimation,handGes2012} start to emerge. Being a depth sensor, the
Kinect camera is not affected by scene illumination and the color of the
clothes worn by the human subject, making object segmentation an easier task.
The challenges still remain are loose clothing, occlusions, and variations in
the style and speed of actions.

In this context, some algorithms have exploited silhouette and edge pixels as
discriminative features. For example, Li et al.\ \cite{Bag3DPoints} sampled
boundary pixels from 2D silhouettes as a bag of features. Yang et
al.\ \cite{DMM} added temporal derivative of 2D projections to get Depth
Motion Maps (DMM).  Vieira et al.\ \cite{STOP} computed silhouettes in 3D by
using the space-time occupancy patterns. Instead of these very simple
occupancy features, Wang et al.\ \cite{Wang2012} computed a vector of 8 Haar
features on a uniform grid in the 4D volume. LDA was used to detect the
discriminative feature positions and an SVM classifier was used for action
classification.  Xia and Aggarwal \cite{DSTIP} proposed an algorithm to
extract Space Time Interest Points (STIPs) from depth sequences and modelled
local 3D depth cuboid using the Depth Cuboid Similarity Feature (DCSF).
However, the accuracy of this algorithm is dependent on the noise level of
depth images. Tang et al.\ \cite{HONV} proposed to use histograms of the
normal vectors computed from depth images for object recognition. Given a
depth image, they computed the spatial derivatives and transformed to polar
coordinates $(\theta, \phi, r)$ where the 2D histograms of $(\theta,\phi)$
were used as object descriptors. Oreifej and Liu~\cite{HON4D} extended these
derivatives to the temporal dimension.  They normalized the gradient vectors
to unit magnitude and projected them onto a fixed basis before
histogramming. In their formulation, the last components of the normalized
gradient vectors were the inverse of the gradient magnitude. As a result,
information from very strong derivative locations, such as edges and
silhouettes, may get suppressed.

In~\cite{ACCV10}, the HOG3D feature is computed within a small 3D cuboid
centered at a space-time point and Sparse Coding (SC) is utilized to obtain a
more discriminative representation. The depth images have a high level of
noise and two subjects may perform one action in different styles. So, to
favor sparsity, SC might select quite different elements in the dictionary
for similar actions (details in Section ~\ref{SC}).  To overcome this problem,
we use the Locality-constrained Linear Coding (LLC)~\cite{LLC} in this paper
to make locality more essential than sparsity.  Moreover, instead of a small
3D cuboid, we divide the input video sequence into subsequences, blocks, and
then cells (Fig.~\ref{fig:overlappedblock}).  The HOG3D features computed at
the cell level are concatenated to form the descriptor at the block
level. Given $m$ classes, the LLC followed by maximum pooling and a {\it
  logistic regression} classifier with {\it L2 regularization} finally give
$m$ probability values for each subsequence. The video sequence is represented
by the concatenation of the subsequence descriptors.  Finally, we use an SVM
for action classification.

We evaluate the proposed algorithm on three standard depth datasets
~\cite{HON4D,handGes2012,Bag3DPoints} and two standard color datasets
~\cite{Weizmann,UCFSports}. We compare the proposed method with ten
state-of-the-art methods~\cite{Wang2012,ActionLet2012,HON4D,3DGrad,
  DSTIP,SDPM,Figure-Centric,mid-level,Hough-Voting, MyWACV14}. Our
experimental results show that our algorithm outperforms these ten methods.

\subsection{Proposed Algorithm}
\label{sec:proposed_alg}
We consider an action as a function operating on a three dimensional space
with $(x,y,t)$ being independent variables and the depth ($d$) being
the dependent variable, i.e., $d = H(x, y, t)$. The discrimination of a
particular action can be characterized by using the variations of the
depth values along these dimensions.

\subsection{Feature Extraction}
\begin{figure}
\begin{center}
 \includegraphics[width=8.3 cm]{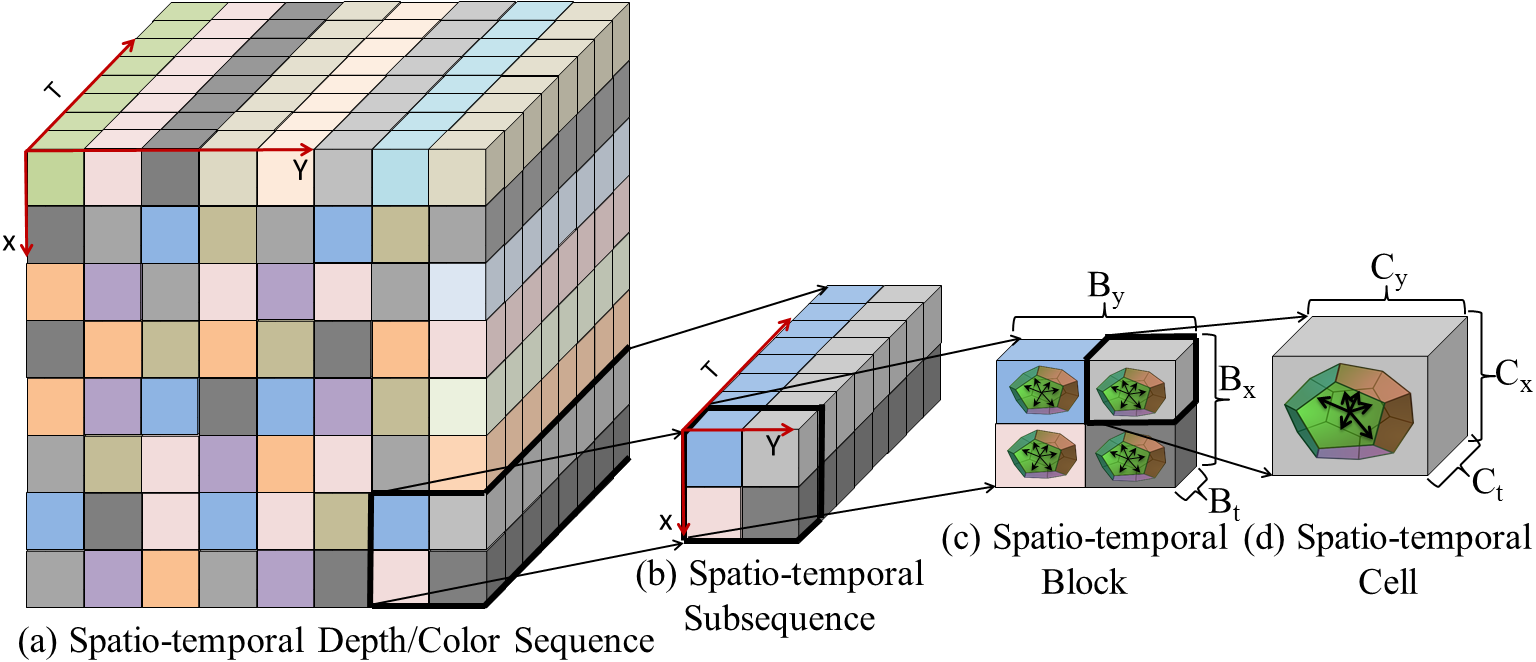}
\end{center}
\caption{(a) An input depth sequence. (b) A spatio-temporal subsequence. (c) A
  spatio-temporal block. (d) A spatio-temporal cell.}
\label{fig:overlappedblock}
\end{figure}

In order to capture sufficient discriminative information such
as local motion and appearance characteristics, the depth sequence is divided
into equally spaced overlapping spatio-temporal subsequences of size $B_x \!
\times \! B_y \!\times\! N_F$, where $N_F$ is the number of frames in the
whole video. Each subsequence is divided into blocks of size $B_x \!\times\!
B_y \!\times\!  B_t$, where $B_t \le N_F$.  Each block is further divided into
equally spaced non-overlapping cells of size $C_x \!\times\! C_y \!\times\!
C_t$. This hierarchy is shown in Fig.~\ref{fig:overlappedblock}.

\subsubsection{Cell Descriptor}
In each cell, a 3D histogram of oriented gradients (HOG) feature \cite{3DGrad}
is computed by evaluating a gradient vector at each pixel in that cell:
\begin{equation}\nabla d(x,y,t)=\frac{\partial d}{\partial x}\hat{\bf i}+\frac
{\partial d}{\partial y}\hat{\bf j}+\frac{\partial d}{\partial t}\hat{ \bf
k},\end{equation}
where the derivatives are given by: ${\partial d}/{\partial x}=(d(x,y,t)-
d(x+\delta x,y, t))/\delta x$, ${\partial d}/{\partial y}=(d(x,y,t)-d(x,y
+\delta y, t))/\delta y$, and ${\partial d}/{\partial t}=(d(x,y,t)-d(x,y,
t+\delta t)/\delta t$.

The HOG feature is computed by projecting each gradient vector onto $n$
directions obtained by joining the centers of $n$ faces of a {\it regular
  n-sided polyhedron} with its center.  A {\it regular dodecahedron} is a type
of {\it regular 12-sided polyhedron} that is commonly used to quantize 3D
gradients. It is composed of 12 {\it regular pentagonal} faces and each face
corresponds to a histogram bin.  Let $V \in \mathds{R}^{3 \!\times\! 12}$ be the
matrix of the center positions ${\bold v}_1, {\bold v}_2, \cdots , {\bold v}_{12}$ of
all faces:
\begin{equation}V=[{\bold v}_1,{\bold v}_2,\cdots,{\bold v}_{12}]\end{equation}

\begin{figure*}
\begin{center}
 \includegraphics[width=16 cm]{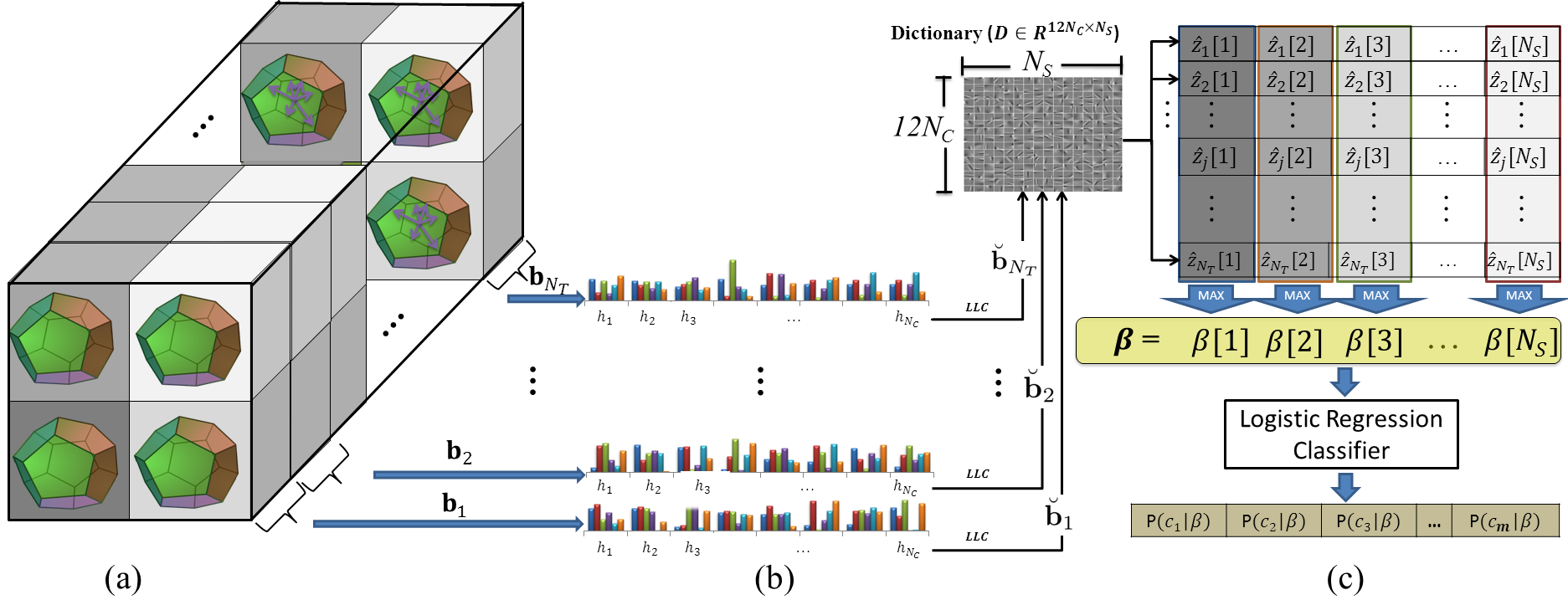}
\end{center}
\caption{The steps for extracting a subsequence descriptor. (a) A
  spatio-temporal subsequence. (b) A block descriptors. (c) A subsequence
  descriptor. }
\label{fig:overallAlgo}
\end{figure*}

For a {\it regular dodecahedron} with center at the origin, these normalized
vectors are given by:
\[{\bold v}_1= \begin{bmatrix}
0 \\[0.3em]
\frac{+1}{L_{\bf v}} \\[0.3em]
\frac{+\varphi}{L_{\bf v}}
\end{bmatrix},
{\bold v}_2= \begin{bmatrix}
0 \\[0.3em]
\frac{-1}{L_{\bf v}} \\[0.3em]
\frac{+\varphi}{L_{\bf v}}
\end{bmatrix},
{\bold v}_3= \begin{bmatrix}
0 \\[0.3em]
\frac{-1}{L_{\bf v}} \\[0.3em]
\frac{-\varphi}{L_{\bf v}}
\end{bmatrix},
{\bold v}_4= \begin{bmatrix}
0 \\[0.3em]
\frac{+1}{L_{\bf v}} \\[0.3em]
\frac{-\varphi}{L_{\bf v}}
\end{bmatrix},
\]
\[{\bold v}_5= \begin{bmatrix}
\frac{+1}{L_{\bf v}} \\[0.3em]
\frac{+\varphi}{L_{\bf v}}\\
0 \\[0.3em]
\end{bmatrix},
{\bold v}_6= \begin{bmatrix}
\frac{-1}{L_{\bf v}} \\[0.3em]
\frac{+\varphi}{L_{\bf v}}\\
0 \\[0.3em]
\end{bmatrix},
{\bold v}_7= \begin{bmatrix}
\frac{-1}{L_{\bf v}} \\[0.3em]
\frac{-\varphi}{L_{\bf v}}\\
0 \\[0.3em]
\end{bmatrix},
{\bold v}_8= \begin{bmatrix}
\frac{+1}{L_{\bf v}} \\[0.3em]
\frac{-\varphi}{L_{\bf v}}\\
0 \\[0.3em]
\end{bmatrix},
\]
\[{\bold v}_9= \begin{bmatrix}
\frac{+\varphi}{L_{\bf v}}\\
0 \\[0.3em]
\frac{+1}{L_{\bf v}} \\[0.3em]
\end{bmatrix},
{\bold v}_{10}= \begin{bmatrix}
\frac{-\varphi}{L_{\bf v}}\\
0 \\[0.3em]
\frac{+1}{L_{\bf v}} \\[0.3em]
\end{bmatrix},
{\bold v}_{11}= \begin{bmatrix}
\frac{-\varphi}{L_{\bf v}}\\
0 \\[0.3em]
\frac{-1}{L_{\bf v}} \\[0.3em]
\end{bmatrix},
{\bold v}_{12}= \begin{bmatrix}
\frac{+\varphi}{L_{\bf v}}\\
0 \\[0.3em]
\frac{-1}{L_{\bf v}} \\[0.3em]
\end{bmatrix},
\]
where $\varphi=\frac{1+\sqrt{5}} {2}$ is the golden ratio, and
$L_{\bf v}=\sqrt{1+\varphi^2}$ is the length of vector ${\bold v}$.
The gradient vector $\nabla d(x,y,t)$ is projected on $V$ to give
\begin{equation}d_V(x,y,t)=\frac {V^\intercal  \nabla d(x,y,t)}
{||\nabla d(x,y,t)||_2} \in \mathds{R}^{12}. \end{equation}

Since $\nabla d(x,y,t)$ should vote into only one single bin in case it is
perfectly aligned with the corresponding axis running through the origin and
the face center, the projected vector $d_V(x,y,t)$ should be quantized. A
threshold value $\psi$ is computed by projecting any two {\it neighboring}
vectors ${\bf v}_i$ and ${\bf v}_j$, i.e.,
\begin{equation}\psi={{\bf v}_i}^\intercal {\bf v}_j=
  \frac{\varphi}{L^2_{\bf v}}.
\end{equation}

The quantized vector is given by
\[ \hat{d}_{V_i}(x,y,t) = \left\{
  \begin{array}{l l}
    0 & \quad \text{if $d_{V_i}(x,y,t) \le \psi$}\\
    d_{V_i}(x,y,t)-\psi & \quad \text{otherwise},
  \end{array} \right.\]
where $1 \leq i \leq 12$. We define ${\bf q}(x,y,t)$ to be $\hat{d}_V(x,y,t)$
scaled by the gradient magnitude, i.e.,
\begin{equation}
  {\bf q}(x,y,t)=\frac {||\nabla d(x,y,t)||_2 \cdot \hat{d_V} (x,y,t)}
  {||\hat{d_V}(x,y,t)||_2},
\end{equation}
and a histogram, ${\bf h} \in \mathds{R}^{12}$, for each cell ${\cal C}$ is
computed: 
\begin{equation}{\bf h}=\frac{1}{C_x C_y C_t}\sum_{(x, y, t) \in
    {\cal C}} {\bf q}(x,y,t).
\end{equation}
Note that this summation is equivalent to histogramming, because each ${\bf
   q}$ vector represents votes in the corresponding bins defined by the 12
 directions of the {\it regular dodecahedron}.

\subsubsection{Block Descriptor}
We combine a fixed number of cells in each
neighborhood into blocks. To get the block descriptor, we vertically
concatenate the histograms $\{{\bf h}_j | j=1,\cdots,N_C\}$ of all cells in
  that block (Fig.~\ref{fig:overallAlgo}(b)):
\begin{equation}
  {\bold b}=[{\bf h}_{1}^\intercal,{\bf h}_{2}^\intercal,\cdots,{\bf h}_{N_C}^
    \intercal]^\intercal \in \mathds{R}^{12 N_C}
\label{eq:blockdescriptor}
\end{equation}
where $N_C= B_xB_yB_t/(C_xC_yC_t)$ denotes the total number of cells in each
block.

The block descriptor ${\bold b}$ in \eqref{eq:blockdescriptor} is normalized
$\hat{{\bold b}} = {\bold b}/{||{\bold b}||_2} \in \mathds{R}^{12
    N_C}$ and a {\it Symmetric Sigmoid} function $f$ is applied to it as a trade-off
between the gradient magnitude and gradient orientation:
\begin{align}
 {\tilde {\bold b}}[k] &= f{(\hat{{\bold b}}[k])}=
 \frac{1-e^{-a \hat{{\bold b}}[k]}}{1+e^{-a \hat{{\bold b}}[k]}} ,
\end{align}
where $k$ is an index: $ 1 \!\leq\! k \!\leq\! 12N_C$. Finally, ${\bold
  {\tilde b}}$ is normalized to give
\begin{equation}
  \breve{{\bold b}}=\frac{{\bold {\tilde b}}}
  {||{\bold {\tilde b}}||_2} \in \mathds{R}^{12 N_C} .
\end{equation}

\subsubsection{Subsequence Descriptor Using SC}
\label{SC}
To compare the descriptors computed using SC and LLC, we show
below how we obtain the intermediate variable $\hat{Z}$ (see
Fig.~\ref{fig:overallAlgo}(c)) using both methods to get two different
representations for the spatio-temporal subsequences.

For each spatio-temporal subsequence, we combine all the spatio-temporal
blocks along the time dimension (see Fig.~\ref{fig:overlappedblock}(b) and
Fig.~\ref{fig:overallAlgo}(a)).  We horizontally concatenate all the block
descriptors within the subsequence to give
\begin{equation}
  A=[\breve{{\bold b}}_1,\breve{{\bold b}}_2,\cdots, \breve{{\bold
        b}}_{N_T}] \in \mathds{R}^{12 N_C \!\times\! N_T} ,
\end{equation}
where $N_T$ denotes the number of blocks within the subsequence.  In the case
where $B_t=1$, then $C_t=1$ also and variable $N_T$ becomes the number of
frames in the depth video. i.e., $N_T=N_F$.

We then create a dictionary $D=[{\bf e}_1, {\bf e}_2, \cdots, {\bf e}_{N_S}]
\in \mathds{R}^{12N_C\!\times\! N_S}$ by applying the $k$-means clustering
algorithm on the computed HOG3D of the training blocks. Here $N_S$ denotes the
number of block descriptors, each has $12N_C$ dimensions. The SC sparsely
encodes each block descriptor $\hat{{\bold b}}^s_i \in A$ into a linear
combination of a few atoms of dictionary $D$ by optimizing:
\begin{equation}
  \hat{Z}\equiv \underset{Z \in \mathds{R}^{N_S \!\times\! N_T}}
  {\operatorname{argmin}} \frac{1}{2}||A-DZ||_2^2 + \lambda||Z||_1 ,
  \label{eq:SC}
\end{equation}
where $\lambda$ is a regularization parameter which determines the sparsity of
the representation of each local spatio-temporal feature.  The twofold
optimization aims at minimizing the reconstruction error and the sparsity of
the coefficient set $\hat{Z} \in \mathds{R}^{N_S \!\times\! N_T}$
simultaneously.

\begin{figure}
\begin{center}
 \includegraphics[width=8.3 cm]{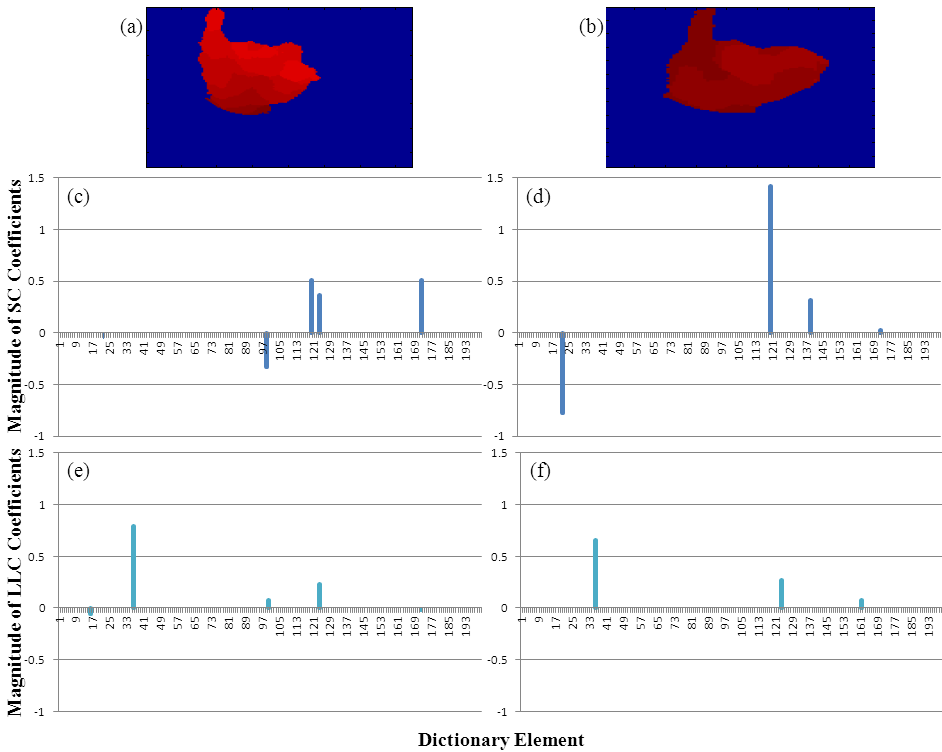}
\end{center}
\caption{(a)-(b) Two blocks at the same spatial location in two videos where
  the same action was performed. (c)-(d) The SC coefficients of these two
  blocks ($||.||_2=1.38$). (e)-(f) The LLC coefficients of these two blocks
  ($||.||_2=0.2$).}
\label{fig:SCLLC}
\end{figure}

\subsubsection{Subsequence Descriptor Using LLC}
An alternative to obtain the coefficient set $\hat{Z} \in \mathds{R}^{N_S
  \!\times\! N_T}$ is to use the LLC. Instead of Eq.~\eqref{eq:SC}, the LLC code
uses the following criteria:
\begin{equation}
  \hat{Z}\equiv \underset{Z \in \Re^{N_S \!\times\! N_T}}
  {\operatorname{argmin}}\frac{1}{2}||A-DZ||_2^2 + \lambda||R \odot Z||_2^2 ,
  \label{eq:LLC}
\end{equation}
where $\odot$ denotes the element-wise multiplication, and $R \in
\mathds{R}^{N_S\!\times\! N_T}$ is the locality adapter that gives different
weights to the basis vectors depending on how similar they are to the input
descriptor $A$. Specifically,
\[ R \!=\! {\small
  \begin{bmatrix}
    \exp (||\breve{{\bold b}}_1-{\bf e}_1||_2 / \sigma) &\cdots & \exp (||\breve{{\bold b}}_{N_T}-{\bf e}_1||_2 / \sigma) \\
    \exp (||\breve{{\bold b}}_1-{\bf e}_2||_2 / \sigma) & \cdots & \exp (||\breve{{\bold b}}_{N_T}-{\bf e}_2||_2 / \sigma) \\
    \cdots & \cdots & \cdots \\
    \exp (||\breve{{\bold b}}_1-{\bf e}_{N_S}||_2 / \sigma) & \cdots & \exp (||\breve{{\bold b}}_{N_T}-{\bf e}_{N_S}||_2 / \sigma) \\
\end{bmatrix}}, \]
 where $\sigma$ is used for adjusting the weight decay speed for the
locality adapter.

For the encoding of block descriptors, we found that locality is more
essential than sparsity, as locality must lead to sparsity but not necessary
vice versa~\cite{LLC}.  So, the LLC should be more discriminative than SC. We
illustrate the superiority of LLC over SC by an example.
Fig.~\ref{fig:SCLLC}(a) shows a block within a spatio-temporal subsequence of
a subject who performed an action and Fig.~\ref{fig:SCLLC}(b) shows the block
in the subsequence at the same spatial location in another video of another
subject doing the same action.  However, the coefficients of SC for these two
blocks (Fig.~\ref{fig:SCLLC}(c)) are quite different (Euclidean distance is
1.38) which means using SC may lead to failure to identify the actions being
the same. On the other hand, the LLC coefficients for these two blocks
(Fig.~\ref{fig:SCLLC}(d)) are very similar to each other (both are distinct
from the SC coefficients) and have a smaller distance apart (Euclidean
distance is 0.2).  As shown in this figure, LLC appears to better represent
the spatio-temporal blocks.

After the optimization step (using Eq.~\eqref{eq:SC} or~\eqref{eq:LLC}), we
get a set of sparse codes $\hat{Z}= [\hat{z}_{1},\cdots,\hat{z}_{N_T}]$, where
each vector $\hat{z}_{i}$ has only a few nonzero elements. It can also be
interpreted that each block descriptor only responds to a small subset of
dictionary atoms.

To capture the global statistics of the subsequence, we use a maximum pooling
function given by ${\boldsymbol \beta}=\zeta(\hat{Z})$
(Fig.~\ref{fig:overallAlgo} (c)), where $\zeta$ returns a vector ${\boldsymbol
  \beta} \in \mathds{R}^{N_S}$ with the $k^{\text{th}}$ element defined as:
\begin{equation}
  \beta[k]=\underset{1 \leq j \leq N_T}{\operatorname{max}}(|\hat{z}_{j}[k]|).
\end{equation}

Through maximum pooling, the obtained vector ${\boldsymbol \beta}$ is the
subsequence descriptor (Fig.~\ref{fig:overallAlgo}(c)).

\subsection{Sequence Descriptor and Classification}
The probability of each action class $c_i$ for a subsequence descriptor
${\boldsymbol \beta}$ is $P(c_i|{\boldsymbol \beta})$. To calculate this
probability, we use a {\it logistic regression} classifier with {\it L2
  regularization} \cite{LR2}. The {\it logistic regression} model was
initially proposed for binary classification; however, it was later on
extended to multi-class classification. Given data $x$, weight $W$, class
label $y$, and a bias term $b$, it assumes the following probability model:
\begin{equation}
  P(y=\pm 1|x,W)=\frac{1}{1+e^{(-y(W^{\intercal} x+b))}}.
\end{equation} 
If the training instances are $x_i$ with labels $y_i \in \{1,-1\}$, for $i =
1, \cdots,l$, one estimates $(W, b)$ by minimizing the
following equation:
\begin{equation*}
  \min_{W} f(W) \equiv \min_{W} \left
    (\frac{1}{2}W^\intercal W+\gamma \sum_{i=1}^{l} \log{(1+e^{-y_iW^
        \intercal x_i})} \right )
\end{equation*}
\begin{equation}
  \text{with } x_i^\intercal\leftarrow [x_i^\intercal,1] \text{ and }
  W^\intercal \leftarrow [W^\intercal,b],
  \label{eq:logistic}
\end{equation}
where $\frac{1}{2}W^\intercal W$ denotes the
{\it L2 regularization} term, and $\gamma>0$ is a user-defined parameter that
weights the relative importance of the two terms. The one-vs-the-rest strategy
by Crammer and Singer \cite{LR1} is used to solve the optimization for this
multi-class problem.

For all the subsequences in the training data corresponding to the same spatial
location $p$, we train one logistic regression classifier. If there are $N_P$
spatial locations (i.e., the total number of overlapping subsequences), we
will have $N_P$ classifiers.  For the classifier corresponding to each spatial
location $p$, we can compute the probability of each action class:
$[P(c_1|{\boldsymbol \beta}_p), P(c_2|{\boldsymbol \beta}_p), \cdots,
P(c_m|{\boldsymbol \beta}_p)]$, where $m$ is the number of class labels.

For an input video sequence, the class probabilities of all the $N_P$
spatio-temporal subsequences are concatenated to create the sequence
descriptor: $[P(c_1|{\boldsymbol \beta}_1),\cdots, P(c_m|{\boldsymbol
  \beta}_1), \cdots, P(c_1|{\boldsymbol \beta}_p),\cdots, P(c_m|{\boldsymbol
  \beta}_p), \cdots \linebreak[4], $ $P(c_1|{\boldsymbol \beta}_{N_P}),
\cdots, P(c_m|{\boldsymbol \beta}_{N_P})]$. In the training stage, the
sequence descriptors of all the training action videos are used to train an
SVM classifier.

In the testing stage, the sequence descriptor of the test video sequence is
computed and fed to the trained classifier.  The class label $c_*$ is defined
to be the one that corresponds to the maximum probability.

\section{Experimental Results}
We evaluate the performance of the proposed algorithm on two different
types of videos: (1) depth and (2) color. For depth data, we use three
standard datasets including the MSRAction3D~\cite{Bag3DPoints,ActionLet2012},
MSRGesture3D~\cite{handGes2012,Wang2012}, and MSRActionPairs3D~\cite{HON4D}. 
For color data, we use two standard datasets including Weizmann
~\cite{Weizmann}, and UCFSports~\cite{UCFSports}.
The performance of our proposed algorithm
is compared with ten state-of-the-art algorithms~\cite{Wang2012,
ActionLet2012,HON4D,3DGrad,DSTIP,SDPM,Figure-Centric,mid-level,Hough-Voting, MyWACV14}.
Except for ~\cite{SDPM,3DGrad}, all accuracies are reported from the original
papers and the codes are obtained from the original authors.

{\bf Datasets.} The MSRGesture3D dataset contains 12 American Sign Language
(ASL) gestures.  Each gesture is performed 2 or 3 times by each of the 10
subjects.  In total, the dataset contains 333 depth sequences.  The
MSRAction3D dataset consists of 567 depth sequences of 20 human sports
actions. Each action is performed by 10 subjects 2 or 3 times.  These actions
were chosen in the context of interactions with game consoles and cover a
variety of movements related to torso, legs, arms and their combinations.  The
MSRActionPairs3D dataset contains 6 pairs of actions, such that within each
pair the motion and the shape cues are similar, but their correlations
vary. For example, {\it Pick up} and {\it Put down} actions have similar
motion and shape; however, the co-occurrence of the object shape and the hand
motion is in different spatio-temporal order. Each action is performed 3 times
using 10 subjects.  In total, the dataset contains 360 depth sequences.

The Weizmann dataset contains 90 video sequences of 9 subjects, each performing
10 actions. The UCFSports dataset consists of videos from sports broadcasts,
with a total of 150 videos from 10 action classes.
Videos are captured in realistic scenarios with complex and
cluttered background, and actions exhibit significant intra-class variation.

{\bf Experimental Settings.} In all the experiments, the ROIs of the depth/RGB
videos are resized to $48 \!\times\!  64$ pixels for the sake of
computation. The size of each cell is $8 \!\times\!  8$ pixels (i.e., $C_x
\!=\! C_y \!=\! 8$) and the number of cells in each block is $2 \!\times\! 2$
along the $X$ and $Y$ dimensions (i.e., $B_x \!=\!B_y\!=\!16$), respectively.
Variables $B_t$ and $C_t$ are set to 1; variable $N_S$, the size of the
dictionary, is set to 200. For RGB videos, we consider the variable $d$
mentioned in Section~\ref{sec:proposed_alg} as the gray-level value of each
pixel.

\subsection{Depth Videos}
In the first experiment, we evaluate the proposed method on the
depth datasets (MSRGesture3D, MSRAction3D and MSRActionPairs3D). To compare
our method with previous techniques on MSRGesture3D dataset, we use
leave-one-subject-out cross-validation scheme proposed by~\cite{Wang2012}. The
average accuracies of our algorithm using SC and LLC are 89.6\% and 94.1\%,
respectively.  These results prove that LLC is more discriminative than SC.
Our algorithm outperformed existing state-of-the-art algorithms (Table
\ref{tab:Depth}). Note that Actionlet method~\cite{ActionLet2012} cannot be
applied to this dataset because of the absence of 3D joint positions.

For the MSRAction3D dataset, we performed experiments, same as previous
works~\cite{ActionLet2012,HON4D}, using five subjects for training
and five subjects for testing. The accuracy obtained is 90.9\% which
is higher than the accuracy 87.1\% from SC, 88.9\% from HON4D~\cite{HON4D} and 89.3\%
from DSTIP~\cite{DSTIP} (Table \ref{tab:Depth}).
For most of the actions, our method achieves near perfect recognition
accuracy. The classification errors occur if two actions are too
similar to each other, such as {\it hand catch} and {\it high throw}.

For the MSRActionPairs3D dataset, same as previous work~\cite{HON4D}, we use
half of the subjects for training and the rest for testing.  The proposed
algorithm has achieved an average accuracy of 98.3\% which is higher than the
accuracy of SC, HON4D and Actionlet methods (Table \ref{tab:Depth}). For nine
of the actions, our method achieves 100\% recognition accuracy; for the
remaining three actions, our method achieves 93\% recognition accuracy.

\begin{table} \large
\centering \caption{Comparison with previous works on three depth datasets:
 MSRGesture3D, MSRAction3D and MSRActionPairs3D.}
    \begin{tabular}{lcccccc}
    \toprule
    {Method} & {Gesture}  & {Action}  & {A/Pairs}\\
\hline \hline
    HOG3D~\cite{3DGrad} & 85.2 & 81.4 & 88.2\\
    ROP~\cite{Wang2012} & 88.5 & 86.5 & -\\
    Actionlet~\cite{ActionLet2012} & NA & 88.2 & 82.2 \\
    HON4D~\cite{HON4D} & 92.4 & 88.9 & 96.7 \\
    DSTIP~\cite{DSTIP} & - & 89.3 & - \\
    RDF~\cite{MyWACV14} & 92.8 & 88.8 & -\\
    \hline
    Our Method (SC) & 89.6 & 87.1 & 94.2 \\
    Our Method (LLC) & {\bf 94.1} & {\bf 90.9} & {\bf 98.3} \\
    \bottomrule
    \end{tabular}
  \label{tab:Depth}
\end{table}

\subsection{RGB Videos}
In the second experiment, we evaluate the proposed method on the
RGB datasets (Weizmann and UCFSports). For Weizmann dataset, we follow the
experimental methodology from~\cite{Weizmann}. Testing is performed by
leave-one-out (LOO) on a per person basis, i.e., for each fold, training is
done on 8 subjects and testing on all video sequences of the remaining
subjects. Our method achieves 100\% recognition accuracy, which is higher than
the accuracy 84.3\% from HOG3D~\cite{3DGrad} based method (Table
\ref{tab:RGB}).  For the UCFSports dataset, most of the reported results used
LOO cross validation and the best result is 91.3\% ~\cite{AFMKL}. The accuracy
of our algorithm with LOO is 93.6\% which is higher than ~\cite{AFMKL}. But,
there are strong scene correlations among videos in certain classes; many
videos are captured in exactly the same location. Lan et
al. ~\cite{Figure-Centric} show that, with LOO, the learning method can
exploit this correlation and memorize the background instead of learning the
action. So, to help alleviate this problem, same as the recently published
works~\cite{mid-level, Figure-Centric, SDPM}, we split the dataset by taking
one third of the videos from each action category to form the test set, and
the rest of the videos are used for training. This reduces the chances of
videos in the test set sharing the same scene with videos in the training
set. The accuracy of our algorithm using this protocol is 93.6\%. Our
algorithm outperformed existing state-of-the-art algorithms (Table
\ref{tab:RGB}) with large margin.  Note that ~\cite{mid-level, Figure-Centric,
  SDPM} used the raw videos whereas our method uses the ROI of the videos. For
fair comparison, we use the ROI of the videos and run the code of ~\cite{SDPM}
that obtained from the original author. Also, ~\cite{Hough-Voting} used
different protocol and split the UCFSports dataset by taking one fifth of the
videos from each action category to form the test set, and the rest of the
videos were used for training.

\begin{table} \large
\centering \caption{Comparison with previous works on two RGB datasets:
 Weizmann and UCFSports.}
    \begin{tabular}{lcccccc}
    \toprule
    {Method} & {Weizmann}  & {UCFSports}\\
    \hline \hline
    HOG3D~\cite{3DGrad} & 84.3 & 76.2 \\
    Hough-Voting~\cite{Hough-Voting} & 97.8 & 86.6\\
    Sparse-Coding~\cite{ACCV10} & - & 84.33\\
    Figure-Centric~\cite{Figure-Centric} & - & 73.1 \\
    Action-Parts~\cite{mid-level} & - & 79.4\\
    SDPM~\cite{SDPM} & {\bf 100} & 79.8\\
    \hline
    Our Method (SC) & {\bf 100} & 87.4 \\
    Our Method (LLC) & {\bf 100} & {\bf 93.6} \\
    \bottomrule
    \end{tabular}
  \label{tab:RGB}
\end{table}

The experimental results show that the proposed method can be applied on color
and depth videos and the accuracies obtained are higher than the
state-of-the-art methods.  Also, the results prove that LLC is more
discriminative than SC for human action recognition problem.

\begin{figure}
\begin{center}
 \includegraphics[width=8.3 cm]{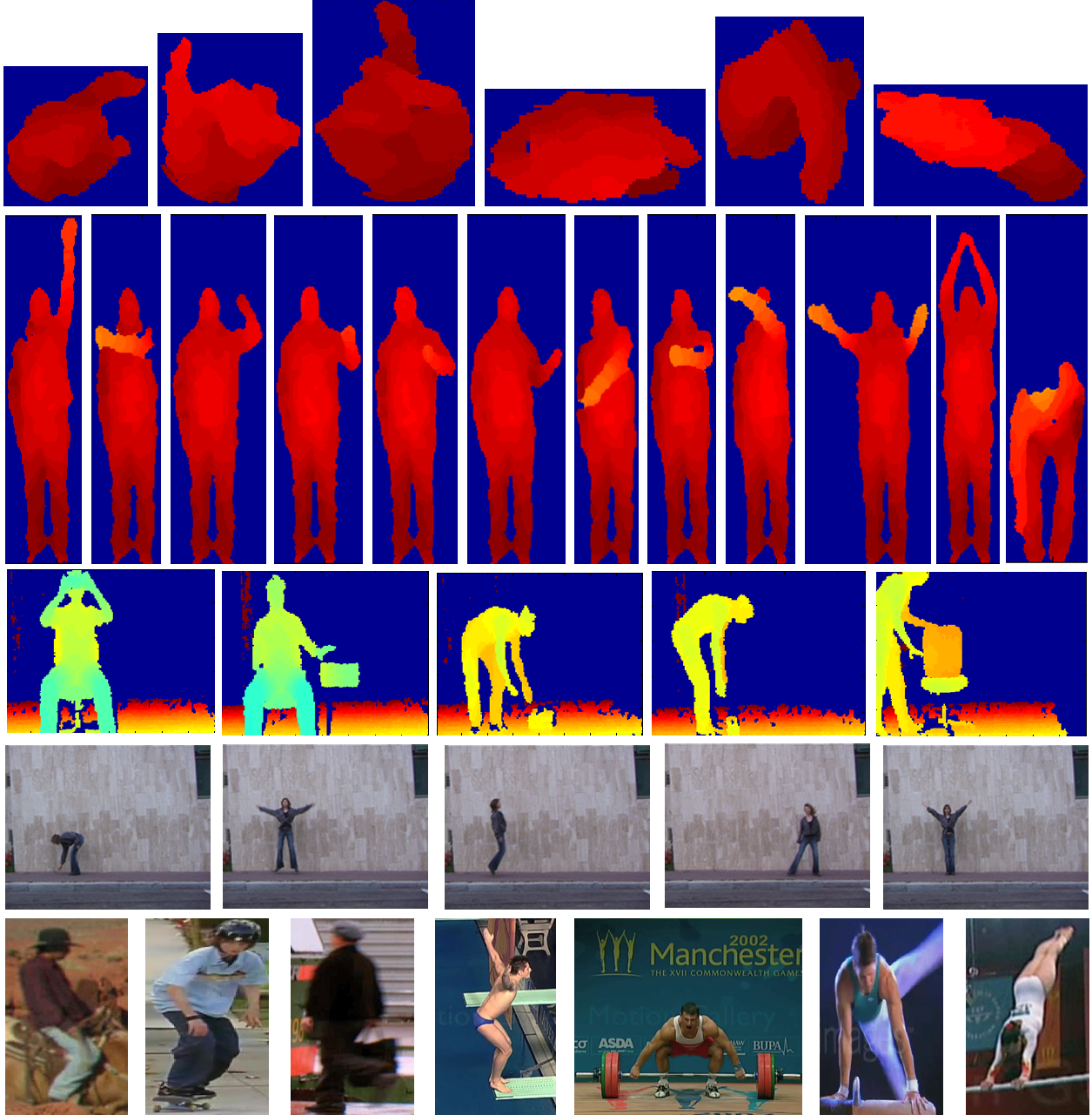}
\end{center}
   \caption{Row-wise from top: Sample depth images from the MSRGesture3D, MSRAction3D,
   MSRActionPairs3D, Weizman, and UCFSports datasets.}
\label{fig:allSamples}
\end{figure}

\section{Conclusion and Future Work}
In this paper, we propose a new action recognition algorithm
which uses LLC to capture discriminative information of human body variation
in each spatio-temporal subsequence of the input depth video.  The proposed
algorithm has been tested and compared with ten state-of-the-art algorithms on
three benchmark depth datasets and two benchmark color datasets. On average,
our algorithm is found to be more accurate than these ten algorithms.

\section*{Acknowledgment}
This research was supported by ARC Grant DP110102399. We thank the the authors of the depth based action classification algorithms~\cite{Wang2012, ActionLet2012,HON4D,DSTIP} for providing the datasets and making their
code publicly available.



%
%
%

\bibliographystyle{IEEEtran}
\balance
\bibliography{bibICPR14}

\begin{thebibliography}{10}
\providecommand{\url}[1]{#1}
\csname url@samestyle\endcsname
\providecommand{\newblock}{\relax}
\providecommand{\bibinfo}[2]{#2}
\providecommand{\BIBentrySTDinterwordspacing}{\spaceskip=0pt\relax}
\providecommand{\BIBentryALTinterwordstretchfactor}{4}
\providecommand{\BIBentryALTinterwordspacing}{\spaceskip=\fontdimen2\font plus
\BIBentryALTinterwordstretchfactor\fontdimen3\font minus
  \fontdimen4\font\relax}
\providecommand{\BIBforeignlanguage}[2]{{%
\expandafter\ifx\csname l@#1\endcsname\relax
\typeout{** WARNING: IEEEtran.bst: No hyphenation pattern has been}%
\typeout{** loaded for the language `#1'. Using the pattern for}%
\typeout{** the default language instead.}%
\else
\language=\csname l@#1\endcsname
\fi
#2}}
\providecommand{\BIBdecl}{\relax}
\BIBdecl

\bibitem{main}
D.~Weinland, M.~\"{O}zuysal, and P.~Fua, ``Making action recognition robust to
  occlusions and viewpoint changes,'' in \emph{ECCV}, 2010.

\bibitem{RGB1}
I.~Everts, J.~V. Gemert, and T.~Gevers, ``Evaluation of color stips for human
  action recognition,'' in \emph{CVPR}, 2013.

\bibitem{RGB2}
F.~Shi, E.~Petriu, and R.~Laganiere, ``Sampling strategies for real-time action
  recognition,'' in \emph{CVPR}, 2013.

\bibitem{MyWACV14}
H.~Rahmani, A.~Mahmood, A.~Mian, and D.~Huynh, ``Real time action recognition
  using histograms of depth gradients and random decision forests,'' in
  \emph{WACV}, 2014.

\bibitem{DSTIP}
L.~Xia and J.~Aggarwal, ``Spatio-temporal depth cuboid similarity feature for
  activity recongition using depth camera,'' in \emph{CVPR}, 2013.

\bibitem{Wang2012}
J.~Wang, Z.~Liu, J.~Chorowski, Z.~Chen, and Y.~Wu, ``Robust {3D} action
  recognition with random occupancy patterns,'' in \emph{ECCV}, 2012, pp.
  872--885.

\bibitem{HON4D}
O.~Oreifej and Z.~Liu, ``{HON4D}: {H}istogram of oriented {4D} normals for
  activity recognition from depth sequences,'' in \emph{CVPR}, 2013.

\bibitem{ActionLet2012}
J.~Wang, Z.~Liu, Y.~Wu, and J.~Yuan, ``Mining actionlet ensemble for action
  recognition with depth cameras,'' in \emph{CVPR}, 2012, pp. 1290 --1297.

\bibitem{DMM}
X.~Yang, C.~Zhang, and Y.~Tian, ``Recognizing actions using depth motion
  maps-based histograms of oriented gradients,'' in \emph{ACM ICM}, 2012.

\bibitem{Bag3DPoints}
W.~Li, Z.~Zhang, and Z.~Liu, ``Action recognition based on a bag of {3D}
  points,'' in \emph{CVPRW}, 2010.

\bibitem{Hand-Pose-Estimation}
C.~Keskin, F.~Kirac, Y.~Kara, and L.~Akarun, ``Real time hand pose estimation
  using depth sensors,'' in \emph{ICCVW}, 2011.

\bibitem{handGes2012}
A.~Kurakin, Z.~Zhang, and Z.~Liu, ``A real time system for dynamic hand gesture
  recognition with a depth sensor,'' in \emph{EUSIPCO}, 2012.

\bibitem{STOP}
A.~W. Vieira, E.~Nascimento, G.~Oliveira, Z.~Liu, and M.~Campos, ``{STOP}:
  {S}pace-time occupancy patterns for {3D} action recognition from depth map
  sequences,'' in \emph{CIARP}, 2012.

\bibitem{HONV}
S.~Tang, X.~Wang, X.~Lv, T.~Han, J.~Keller, Z.~He, M.~Skubic, and S.~Lao,
  ``Histogram of oriented normal vectors for object recognition with a depth
  sensor,'' in \emph{ACCV}, 2012.

\bibitem{ACCV10}
Y.~Zhu, X.~Zhao, Y.~Fu, and Y.~Liu, ``Sparse coding on local spatial-temporal
  volumes for human action recognition,'' in \emph{ACCV}, 2010.

\bibitem{LLC}
J.~Wang, J.~Yang, K.~Yu, F.~Lv, T.~Huang, and Y.~Gong, ``Locality constrained
  linear coding for image classification,'' in \emph{CVPR}, 2010.

\bibitem{Weizmann}
M.~Blank, L.~Gorelick, E.~Shechtman, M.~Irani, and R.~Basri, ``Actions as
  space-time shapes,'' in \emph{ICCV}, 2005, pp. 1395--1402.

\bibitem{UCFSports}
M.~Rodriguez, J.~Ahmed, and M.~Shah, ``Action mach a spatio-temporal maximum
  average correlation height filter for action recognition,'' in \emph{CVPR},
  2008.

\bibitem{3DGrad}
A.~Klaeser, M.~Marszalek, and C.~Schmid, ``A spatio-temporal descriptor based
  on 3d-gradients,'' in \emph{BMVC}, 2008.

\bibitem{SDPM}
Y.~Tian, R.~Sukthankar, and M.~Shah, ``Spatiotemporal deformable part models
  for action detection,'' 2013.

\bibitem{Figure-Centric}
T.~Lan, Y.~Wang, and G.~Mori, ``Discriminative figure-centric models for joint
  action localization and recognition,'' in \emph{ICCV}, 2011.

\bibitem{mid-level}
M.~Raptis, I.~Kokkinos, and S.~Soatto, ``Discovering discriminative action
  parts from mid-level video representations,'' in \emph{CVPR}, 2012.

\bibitem{Hough-Voting}
A.~Yao, J.~Gall, and L.~Gool, ``A hough transform-based voting framework for
  action recognition,'' in \emph{CVPR}, 2010.

\bibitem{LR2}
R.-E. Fan, K.-W. Chang, C.-J. Hsieh, X.-R. Wang, and C.-J. Lin, ``Liblinear: A
  library for large linear classification,'' \emph{Journal of Machine Learning
  Research}, 2008.

\bibitem{LR1}
S.~S. Keerthi, S.~Sundararajan, K.-W. Chang, C.-J. Hsieh, and C.-J. Lin, ``A
  sequential dual method for large scale multi-class linear svms,'' in
  \emph{ACM SIGKDD}, 2008.

\bibitem{AFMKL}
X.~Wu, D.~Xu, L.~Duan, and J.~Luo, ``Scalable action recognition with a
  subspace forest,'' in \emph{CVPR}, 2012.

\end{thebibliography}

\end{document}